\documentclass[11pt]{article}

\usepackage[margin=1in]{geometry}
\usepackage[T1]{fontenc}
\usepackage[utf8]{inputenc}
\usepackage{lmodern}
\usepackage{microtype}
\usepackage{setspace}
\usepackage{amsmath,amssymb,mathtools}
\usepackage{booktabs}
\usepackage{longtable}
\usepackage{array}
\usepackage{multirow}
\usepackage{siunitx}

\usepackage{graphicx}
\usepackage{subcaption}
\usepackage{float}
\usepackage{placeins}
\usepackage{xcolor}

\usepackage{enumitem}
\usepackage[hidelinks]{hyperref}
\hypersetup{
    pdftitle={A Statistical and Machine Learning Framework for Quantifying Offensive Impact in Professional Box Lacrosse},
    pdfauthor={Robert Jimerson Jr.},
    pdfsubject={Expected goals and recorded offensive attribution in professional box lacrosse},
    pdfkeywords={box lacrosse, National Lacrosse League, expected goals, machine learning, offensive attribution}
}
\usepackage[nameinlink,noabbrev]{cleveref}
\usepackage[round,authoryear]{natbib}

\usepackage{listings}
\usepackage{appendix}

\title{A Statistical and Machine Learning Framework for Quantifying Offensive Impact in Professional Box Lacrosse}
\author{Robert Jimerson Jr.\\
\small Independent Researcher}
\date{September 2026}

\begin{document}

\maketitle

\begin{abstract}
Traditional box-lacrosse statistics summarize realized outcomes but say little about shot quality or the different roles involved in creating scoring opportunities. This study develops a documented framework for estimating expected goals (xG) and attributing recorded offensive value in the National Lacrosse League, using 1,006 manually annotated Rochester Knighthawks shot attempts, including 151 goals, from thirteen consecutive games played between January 3 and April 4 of the 2025--2026 season. Goals, saves, misses, and blocked attempts were recorded with spatial, temporal, situational, passing, and two-player-action context.

Logistic regression, random forest, and extremely randomized trees were crossed with three nested feature specifications and evaluated by Leave-One-Game-Out cross-validation against a constant predictor assigning each held-out shot its training-fold goal rate. The contextual baseline random forest had the lowest observed pooled log loss and Brier score, improving on that benchmark by 1.22\% and 1.50\%, respectively, while five of the nine specifications failed to beat it. These are descriptive comparisons from the same cross-validation used for model selection, not an independent post-selection assessment. The recorded context therefore supplies limited probability information beyond the base rate at this sample size. Adding two-man-action role and qualifying pick type did not improve the primary metrics, though a pick-aware model was retained because the baseline cannot express the required no-pick intervention.

The probabilities support attribution through xG for shooters and shot-based expected assists for recorded final passers. Core Offensive Impact combines those two supported components. A counterfactual Expected Pick Value (xPV) for qualifying pickers compares the recorded state with the same row after the pick fields are set to their no-pick values. Its magnitude was indistinguishable from model noise. Its direction exceeded all 200 row-permutation replicates, but that diagnostic has limited tail resolution and does not preserve game-level pick composition; xPV is therefore reported separately and only within an explicitly exploratory augmented score. Given thirteen games involving one focal team, the framework and player summaries should be interpreted as an initial case study and documented baseline rather than league-wide or causal estimates.

\end{abstract}

\section{Introduction}
\label{sec:introduction}

\subsection{Motivation and Problem Context}
\label{subsec:introduction_motivation}

Statistical and machine-learning methods have expanded how offensive performance is evaluated in professional sports. Rather than relying exclusively on final outcomes, modern analytical frameworks can estimate the quality of scoring opportunities and identify the actions contributing to their creation. Expected-goals models have become a prominent example in soccer and ice hockey, where shot-level probabilities supplement traditional measures such as goals and total shots \citep{fairchild2018spatial,macdonald2012weighted}. Comparable publicly documented methods remain less developed in professional box lacrosse.

Box lacrosse presents a valuable setting for opportunity-based analysis. The sport is played at a fast pace on an enclosed surface, with frequent interactions among shooters, passers, and off-ball players. Scoring opportunities often develop through passing sequences, on-ball picks, and off-ball picks, making the shooter only the final participant in an offensive sequence that may involve several teammates. Publicly available National Lacrosse League statistics primarily summarize realized outcomes through measures such as goals, assists, points, and shots on goal \citep{nllpublicstats}. Although these metrics remain important, goals do not account for differences in shot quality, assists depend on conversion by the receiving player, and shots on goal exclude missed and blocked attempts. Traditional box-score statistics also provide little or no credit for a player who creates separation through a pick without touching the ball.

These limitations make it difficult to distinguish opportunity generation from conversion. A player's goal total may reflect shot volume, opportunity quality, finishing, or a combination of these factors. Similarly, assist totals do not identify high-quality opportunities that teammates fail to convert, while valuable off-ball contributions may remain statistically invisible. A more comprehensive evaluation system therefore requires complete shot-attempt data, spatial estimates of shot quality, and a method for assigning expected value to the players directly involved in producing an attempt.

\subsection{Research Objective and Framework Overview}
\label{subsec:introduction_objective}

The primary objective of this study is to develop and evaluate a documented statistical and machine-learning framework for quantifying recorded offensive impact in professional box lacrosse. Methodological transparency here refers to specifying the event definitions, feature construction, estimator settings, random seed, and game-level validation design so that the procedure can be implemented on an independently collected dataset. It does not imply that the present numerical results can be replicated without the unreleased event data and code. The empirical analysis uses a frozen manually annotated dataset containing 1,006 shot attempts from thirteen consecutive Rochester Knighthawks games played between January 3 and April 4 of the 2025--2026 NLL season. These observations include 151 goals together with saved, missed, and blocked attempts. Each shot was supplemented with spatial and contextual information derived from structured video review and processed through an automated feature-engineering and validation pipeline.

Expected goals provide the probabilistic foundation of the framework. Each shot is assigned an xG value representing its estimated probability of becoming a goal. Logistic regression, random forest, and extremely randomized trees were compared using Leave-One-Game-Out cross-validation. Each algorithm was evaluated with a contextual baseline, the baseline plus two-man-action role, and the baseline plus both two-man-action role and qualifying pick type. This design tested whether the pick fields improved probability estimates for games excluded from training.

The resulting probabilities support player attribution through xG and expected assists (xA): the shooter accumulates xG through recorded attempts, and an identified final passer receives the resulting shot value as xA. Core Offensive Impact (OI) combines these supported shooting and final-passing components. Completed picker annotations additionally support an exploratory counterfactual Expected Pick Value (xPV), defined as the difference between a pick-aware model's observed-state probability and its probability after the pick variables are changed to the no-pick state. An exploratory augmented OI adds xPV for descriptive illustration, but xPV is not treated as a validated goal-equivalent quantity. Finishing Above Expected (FAE) remains separate so that opportunity creation is not conflated with realized shooting outcomes. Both cumulative xG and xG per shot are reported to distinguish volume from average modeled opportunity quality.

\subsection{Research Questions and Contributions}
\label{subsec:introduction_contributions}

The study was guided by three research questions:

\begin{enumerate}
    \item \textbf{RQ1:} Which evaluated statistical or machine-learning specification produces the lowest observed held-out probability error for NLL games excluded from model training?

    \item \textbf{RQ2:} Do two-man-action role and qualifying pick type improve held-out probability quality beyond the contextual baseline?

    \item \textbf{RQ3:} How can shot-level expected-goals estimates be extended to attribute recorded offensive value to shooters, final passers, and qualifying pickers while supporting practical player and game analysis?
\end{enumerate}

This study makes three principal contributions. First, it establishes a standardized event representation, video-annotation methodology, and automated preprocessing pipeline for professional box lacrosse. The resulting data include all observed shot outcomes together with spatial, temporal, situational, passing, and two-man-action information.

Second, the study develops and evaluates an NLL-specific expected-goals model using game-level validation and probability-sensitive evaluation measures. The comparison crosses three fixed algorithms with three nested feature sets, each judged against a constant predictor assigning every held-out shot its training-fold goal rate. A contextual baseline random forest produced the lowest observed aggregate out-of-fold log loss and Brier score, improving on that benchmark by 1.22\% and 1.50\%, respectively; four of the nine specifications beat the benchmark and five did not. Because the same cross-validation predictions were used for comparison and model selection, these differences are descriptive rather than an independent estimate of post-selection performance. The recorded shot context therefore supplies limited probability information beyond the base rate at this sample size, and the model is offered as a documented starting point rather than a strong predictor. Adding two-man-action and pick-type variables did not improve the primary probability metrics.

Third, the study extends shot probabilities into an attribution system encompassing xG, xA, core OI, exploratory counterfactual xPV, and FAE, and reports empirical player results from the completed annotation. These measures provide a documented structure for player evaluation, game-level analysis, passing and pick connections, and coaching reports. Because the analysis contains thirteen games involving one team, the model should be interpreted as an initial empirical baseline rather than a definitive league-wide estimate.

The remainder of the paper is organized as follows. Section~\ref{sec:related_work} reviews related research. Sections~\ref{sec:dataset} and~\ref{sec:event_representation} describe the dataset and feature-engineering methodology. Sections~\ref{sec:methodology} and~\ref{sec:model_results} present model development, evaluation, and interpretation. Section~\ref{sec:offensive_impact} defines the Offensive Impact framework, Section~\ref{sec:empirical_oi} reports the empirical attribution results, Section~\ref{sec:applications} describes practical applications, and Sections~\ref{sec:discussion} and~\ref{sec:conclusion} discuss the findings and conclude the paper.

\section{Related Work}
\label{sec:related_work}

\subsection{Expected Goals and Shot-Quality Modeling}
\label{subsec:related_work_xg}

Expected-goals models estimate the probability that an individual shot will result in a goal, conditional on information describing the attempt and its context. The resulting probability represents the expected scoring value of the shot and can be aggregated across players, teams, or games. This approach distinguishes opportunity volume from estimated opportunity quality, providing information that goals and unweighted shot counts do not capture \citep{fairchild2018spatial,mead2023expected}.

Spatial information forms the foundation of many expected-goals models developed for soccer. Shot coordinates are commonly transformed into distance and angle relative to the goal, with scoring probability generally decreasing as distance increases or the shooting position becomes less favorable. Fairchild et al.\ developed a logistic regression model using manually annotated Major League Soccer shots and found that distance and angle provided the principal predictive information within their feature set \citep{fairchild2018spatial}.

More detailed event and tracking data have allowed researchers to incorporate information beyond shot location. Lucey et al.\ analyzed nearly 10,000 shots using player and ball tracking data from the period preceding each attempt. Their model included game phase, defender proximity, interactions among surrounding players, speed of play, and shot location \citep{lucey2015quality}. Subsequent studies have evaluated logistic regression, gradient-boosted trees, neural networks, and other machine-learning methods using combinations of spatial, situational, player-level, and preceding-event features \citep{hewitt2023machine,mead2023expected}.

Expected-goals methodology has also been applied to ice hockey, which shares an enclosed rink environment and several tactical characteristics with professional box lacrosse. Macdonald et al.\ assigned goal probabilities to National Hockey League shots using information such as distance, shot type, game situation, rebound status, and coordinates. The resulting probability-weighted shots were used to evaluate skaters, goaltenders, and teams, demonstrating applications beyond predicting individual shot outcomes \citep{macdonald2012weighted}.

Because xG represents a probability, evaluation must consider the numerical quality of the estimates. Classification accuracy cannot distinguish between models that produce the same binary decision while assigning substantially different probabilities. Probability-sensitive measures, discrimination, and calibration therefore provide complementary evidence about model quality \citep{fairchild2018spatial,mead2023expected}. Collectively, the soccer and hockey literature supports distance and angle as an interpretable spatial baseline while showing that contextual features and flexible algorithms may provide additional value when sufficient representative data are available. Their benefit must nevertheless be demonstrated empirically within the sport being modeled.

\subsection{Opportunity Creation and Player Evaluation}
\label{subsec:related_work_opportunity_creation}

Goals and assists provide important summaries of offensive production, but both depend on relatively rare scoring outcomes and reveal little about the quality of the preceding actions. Players who repeatedly create valuable opportunities may therefore receive limited statistical credit when their teammates fail to convert them. Expected assists address part of this limitation by extending probability-based evaluation from shooters to passers. In one common shot-based formulation, xA credits the final passer with the expected scoring value of the resulting shot, regardless of its outcome \citep{martens2021space}. Implementations differ in whether they evaluate every pass, the probability that a pass will produce a shot, or only passes that directly precede recorded attempts.

Broader action-value frameworks extend this principle beyond shots and final passes. Decroos et al.\ introduced a framework that values soccer actions according to how they change a team's probabilities of scoring and conceding. This approach assigns context-dependent values to passes, dribbles, shots, and other actions, allowing player contributions to be aggregated without restricting evaluation to goals and assists \citep{decroos2019actions}. Expected-possession-value research in basketball has similarly used player-tracking data to estimate how movements and decisions change the expected outcome of a possession \citep{cervone2016multiresolution}.

Tracking data have also enabled the evaluation of contributions made without possession of the ball. Spearman developed an off-ball scoring opportunity model that recognizes attackers who move into valuable scoring positions even when they never receive the ball \citep{spearman2018beyond}. Fernández and Bornn developed related methods for quantifying space occupation and generation, including movements that create space for teammates by drawing defenders elsewhere \citep{fernandez2018wide}. These studies demonstrate that offensive value can arise from positioning and movement that traditional event counts do not capture.

This literature provides the conceptual foundation for separating shooting, passing, and off-ball opportunity creation. However, comprehensive action- and possession-value models frequently require continuous tracking data or complete event sequences. When those data are unavailable, structured video annotation offers a more limited but practical method for identifying observable contributions associated with shots.

\subsection{Lacrosse Analytics and Data Availability}
\label{subsec:related_work_lacrosse}

Published lacrosse analytics research remains comparatively limited, but prior studies have demonstrated the value of shot-level information. Schanzenbach et al.\ analyzed manually recorded Major League Lacrosse attempts using variables such as location, distance, handedness, shot situation, assists, and outcome. Their spatial zones and shot-density representations supported the evaluation of player and team tendencies relative to league averages \citep{schanzenbach2017shotquality}. Myers et al.\ subsequently developed expected-goals and post-shot expected-goals models using 4,497 shots from 55 NCAA Division I men's field-lacrosse games. Their logistic regression model incorporated distance, angle, and whether the attempt followed a pass. Within their sample, the resulting probability-based measures were more strongly correlated with winning percentage than unweighted shots and shots-on-target counts \citep{myers2021lacrossexg}.

Research specific to professional box lacrosse is more limited. Tabone conducted a possession-level analysis of the Rochester Knighthawks during the 2017--2018 NLL season using information that included shot distance, target location, shot-clock timing, possession outcomes, and shots generated per possession \citep{tabone2018boxlacrosse}. Although this work did not establish a probabilistic player-attribution framework, it demonstrated that detailed event logging could support empirical evaluation of NLL offensive strategy.

At the time of data collection, publicly available NLL statistics primarily consisted of conventional box-score measures such as goals, assists, points, shots on goal, loose balls, and turnovers \citep{nllpublicstats}. These data describe observed production but do not constitute a research-ready shot-level dataset containing coordinates, outcomes for every attempt, preceding passes, offensive situations, and qualifying pick actions. This limitation in public accessibility does not imply that the league, its teams, or commercial partners do not possess more detailed proprietary information. Rather, the observations required for independent development and validation were not publicly available in research-ready form.

Models developed for field lacrosse also cannot be transferred directly to professional box lacrosse without validation. The sports differ in playing geometry, goal dimensions, shot-clock rules, personnel configurations, substitution patterns, and offensive structure. The smaller enclosed surface and prominent use of picks and two-man actions also produce a distinct scoring environment. These differences, together with the absence of an accessible event-level NLL dataset, motivated the structured video-annotation methodology described in Section~\ref{sec:dataset}.

\subsection{Research Gap and Positioning}
\label{subsec:related_work_gap}

The literature establishes the value of shot-quality modeling in field lacrosse and detailed event logging in professional box lacrosse. Based on the publicly accessible work reviewed for this study, however, no prior research was identified that combined an NLL-specific expected-goals model with player attribution for shooting, passing, and directly shot-producing pick actions under a common validation framework.

The present study addresses this gap in three ways. First, it creates a standardized shot-level representation from manually annotated NLL game video, including goals, saves, misses, and blocked attempts. Second, it compares statistical and machine-learning models using Leave-One-Game-Out cross-validation, probability-sensitive evaluation measures, calibration analysis, and a base-rate benchmark against which every specification is judged. Third, it extends shot-quality estimates to player attribution: xG measures the value of a player's shots, xA assigns the value of a resulting shot to its recorded final passer, and counterfactual Expected Pick Value compares observed and no-pick scoring probabilities for qualifying on-ball or off-ball actions. Core Offensive Impact combines xG and xA; xPV is reported only in a clearly labeled exploratory augmented score, while finishing above expected remains separate as a measure of shot conversion.

The contribution is therefore not the claim that advanced analysis has never been conducted in box lacrosse. Rather, it is the integration of NLL event collection, sport-specific probability modeling, game-level validation, and on-ball and off-ball player attribution within a single documented framework. Because the empirical analysis contains thirteen games involving one team, the resulting model should be interpreted as an initial case study and baseline rather than a definitive league-wide estimate. The methodology is intended to be recalibrated and extended as additional teams, seasons, and event types are incorporated.

\section{Dataset}
\label{sec:dataset}

\subsection{Data Source and Scope}

Public National Lacrosse League (NLL) statistics report outcomes such as goals, assists, and shots on goal but do not provide the shot-level spatial and contextual information required for expected-goals analysis. To the author's knowledge, no public NLL dataset contained this information when the study began. A custom event-level dataset was therefore created through manual video review.

The frozen dataset covers thirteen consecutive Rochester Knighthawks regular-season games from January 3 through April 4, 2026. These were the NLL+ replays used in the completed annotation window; the sample is not the full eighteen-game Rochester season. It excludes the first two regular-season games (December 14 and December 20, 2025) and the final three (April 11, April 18, and April 19, 2026). Table~\ref{tab:game_inventory} lists the included games, whose Rochester goal totals sum to the 151 goals in the event file. Dates, opponents, locations, and scores were reconciled against the official schedule and results \citep{rochester2026schedule}; broadcast video was accessed through NLL+ \citep{nllplus2026}.

Every Rochester shot attempt in the included broadcasts was annotated with its location, outcome, and observable game context. Rochester was selected as the initial case study because the research originated from a Rochester-specific coaching question and sustained manual collection for one focal team was practically feasible. The team did not sponsor, participate in, supply data for, or review the study. The event definitions and preprocessing procedures were designed for application to additional teams and seasons, and the annotation schema in Section~\ref{sec:event_representation} is treated as versioned and append-only so that subsequent collection can be combined with the present file without redefining existing fields.

\begin{table}[ht]
    \centering
    \footnotesize
    \caption{Rochester Knighthawks games included in the frozen analysis dataset. Scores are shown from Rochester's perspective; OT denotes overtime.}
    \label{tab:game_inventory}
    \begin{tabular}{@{}rlll@{}}
        \toprule
        \textbf{Date} & \textbf{Opponent} & \textbf{Site} & \textbf{Result} \\
        \midrule
        Jan. 3, 2026 & Colorado Mammoth & Away & L, 13--18 \\
        Jan. 9, 2026 & Philadelphia Wings & Away & W, 16--13 \\
        Jan. 10, 2026 & Buffalo Bandits & Home & W, 12--9 \\
        Jan. 17, 2026 & Toronto Rock & Home & L, 10--11 \\
        Jan. 31, 2026 & Vancouver Warriors & Home & L, 6--16 \\
        Feb. 7, 2026 & Vancouver Warriors & Away & L, 15--16 \\
        Feb. 14, 2026 & Ottawa Black Bears & Home & L, 8--10 \\
        Feb. 28, 2026 & Saskatchewan Rush & Home & W, 13--12 (OT) \\
        Mar. 8, 2026 & Calgary Roughnecks & Home & L, 7--14 \\
        Mar. 15, 2026 & Las Vegas Desert Dogs & Away & L, 10--17 \\
        Mar. 21, 2026 & Las Vegas Desert Dogs & Home & L, 13--17 \\
        Mar. 28, 2026 & Oshawa FireWolves & Away & W, 18--11 \\
        Apr. 4, 2026 & Halifax Thunderbirds & Away & L, 10--15 \\
        \bottomrule
    \end{tabular}
\end{table}

\subsection{Data Collection}

Each game was reviewed in full and annotated by the author using the Shot-Plotter application \citep{shotplotter2026}. Each row represents one attempt; goals, saved shots, misses, and blocks were all retained. When broadcast visibility was insufficient to assign a contextual category confidently, the observation was left unassigned rather than manually inferred; fold-specific imputation subsequently handled those values as described in Section~\ref{sec:methodology}. The recorded fields and derived variables are described in Section~\ref{sec:event_representation}.

Quality assurance combined manual and programmatic checks. Goal totals were compared with official NLL game statistics, and all games were reviewed a second time to identify omitted or incorrectly classified attempts. Automated validation compared raw and processed record counts and checked for missing values, invalid ranges, and unexpected categories. Data and preprocessing inconsistencies identified during development were corrected before the final dataset was regenerated and the model experiments were run.

\subsection{Dataset Summary}

The frozen analysis file contains 1,006 attempts: 151 goals, 587 saved shots, 203 misses, and 65 blocks. The observed goal rate was 15.01\%. A final passer was recorded for 881 attempts. The picker annotation identified 225 qualifying directly shot-producing picks, including 210 on-ball and 15 off-ball actions. Because all observations come from one team over thirteen games, the data constitute an initial team-level case study rather than a league-wide sample.

\begin{table}[ht]
    \centering
    \caption{Summary of the Rochester Knighthawks shot dataset.}
    \label{tab:dataset_summary}
    \begin{tabular}{lr}
        \toprule
        \textbf{Characteristic} & \textbf{Value} \\
        \midrule
        Season & 2025--2026 \\
        Team & Rochester Knighthawks \\
        Games & 13 \\
        Shot attempts & 1,006 \\
        Goals & 151 \\
        Saved shots & 587 \\
        Missed shots & 203 \\
        Blocked shots & 65 \\
        Non-goals & 855 \\
        Observed goal rate & 15.01\% \\
        Attempts with final passer & 881 \\
        Qualifying picks & 225 \\
        \bottomrule
    \end{tabular}
\end{table}

The outcome imbalance motivated probability-based evaluation measures rather than classification accuracy alone. Dependence among attempts from the same contest motivated game-level cross-validation (Section~\ref{subsec:crossvalidationstrategy}).

\subsection{Sentinels and Picker Validation}
\label{subsec:picker_dataset}

The games were reviewed to identify the non-shooting picker for qualifying directly shot-producing two-player actions and to classify each action as on-ball or off-ball. A missing passer, disabled shot clock, or absent qualifying picker was encoded as $-1$. Absent picks were paired with the categories ``no qualifying pick'' and ``not two-man action''; qualifying actions were labeled ``on-ball'' or ``off-ball'' and identified the credited picker. Automated checks required picker number, pick type, and two-man-action state to agree. Picker number was retained only for attribution and excluded from every model input. Appendix~\ref{app:feature_dictionary} gives the exact stored field names and values.

\subsection{Coordinate Normalization}

Shot locations were recorded as raw \(x\)- and \(y\)-coordinates in their original court orientation. Within each game and period, the attacked goal was identified from the median horizontal shot coordinate. Periods attacking the left goal were mirrored so every attempt was expressed relative to the right-side goal at $(85,0)$. Raw coordinates and the inferred attacked-goal coordinate were retained for audit; the model used the normalized horizontal coordinate and derived spatial features. After correction and normalization, mean shot distance was 21.18 feet and the maximum was 78.81 feet. Section~\ref{sec:event_representation} gives the geometric definitions.

\begin{figure}[t]
    \centering
    \includegraphics[width=0.85\textwidth]
        {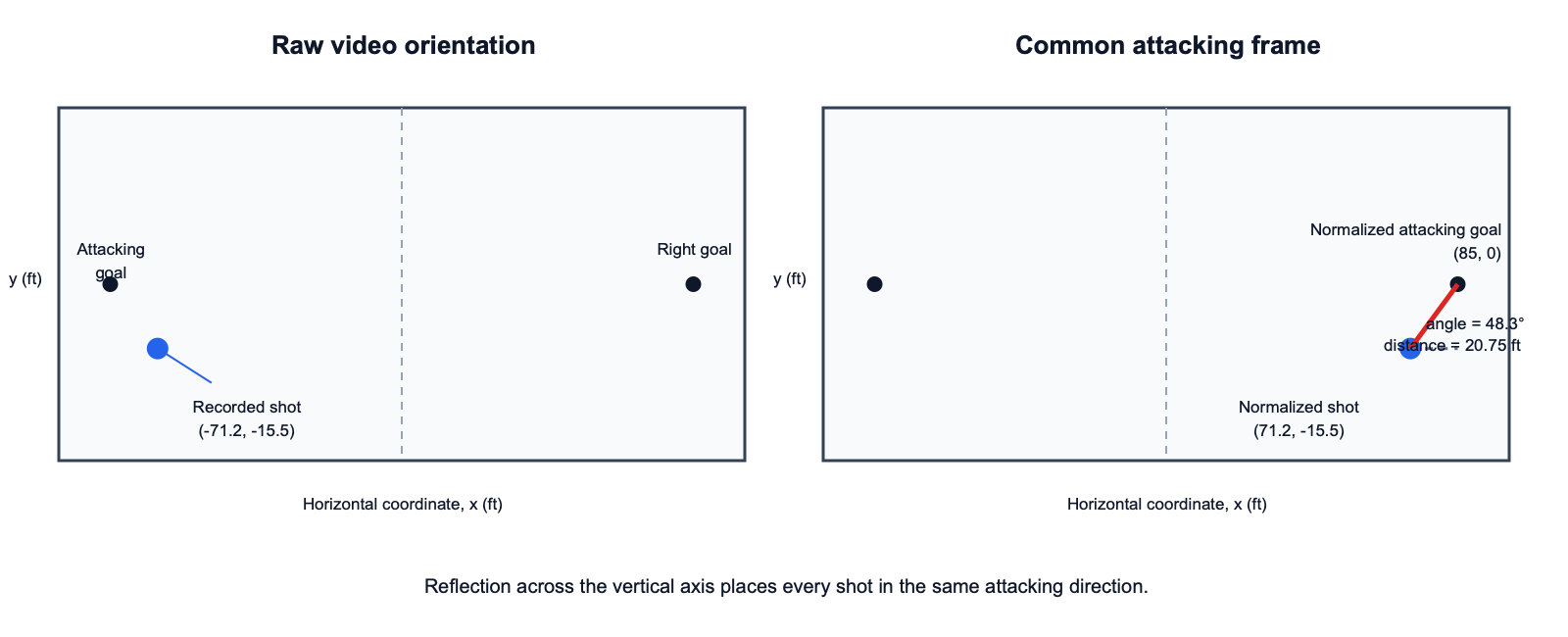}
    \caption{Illustration of coordinate normalization into a common attacking direction before distance and angle are calculated.}
    \label{fig:coordinate_normalization}
\end{figure}

\subsection{Dataset Limitations}

The single-team, thirteen-game sample limits generalization to other teams, playing styles, and seasons. Manual annotation may also introduce measurement or classification error despite repeated review and automated validation. Finally, event video does not provide continuous player tracking, so the data cannot directly measure defensive positioning, spacing, or every off-ball contribution. These limitations motivated game-level validation and conservative interpretation; broader league and tracking data are needed to test generalizability.

\section{Event Representation and Feature Engineering}
\label{sec:event_representation}

\subsection{Event Representation}

Each attempt was represented by ten numeric and seven categorical baseline fields (Table~\ref{tab:event_features}). Two nested extensions added two-man-action role and qualifying pick type. Shooter, passer, and picker identifiers were excluded from the default model inputs; picker number was prohibited from every feature set and used only after prediction for xPV attribution.

\begin{table}[t]
\centering
\small
\caption{Feature groups used to represent each shot attempt.}
\label{tab:event_features}
\begin{tabular}{p{0.22\textwidth}p{0.68\textwidth}}
\toprule
\textbf{Feature group} & \textbf{Variables} \\
\midrule
Baseline numeric & Quarter, time remaining, shot-clock seconds, shot-clock-disabled indicator, passer-present indicator, normalized $X$, $Y$, distance, angle, and behind-goal indicator \\
Baseline categorical & Strength state, offense type, swing state, shot type, off-pass state, score differential, and pulled-goalie state \\
Two-man extension & Baseline plus two-man-action role \\
Pick-type extension & Two-man extension plus qualifying pick type \\
Outcome & Goal, saved, blocked, or missed \\
\bottomrule
\end{tabular}
\end{table}

\subsection{Spatial Features}

After the normalization described in Section~\ref{sec:dataset}, the attacking goal was located at $(85,0)$. For normalized horizontal coordinate $x_i$ and vertical coordinate $y_i$, distance was

\begin{equation}
d_i=\sqrt{(85-x_i)^2+y_i^2},
\label{eq:shot_distance}
\end{equation}

and absolute angle was

\begin{equation}
\theta_i=
\operatorname{atan2}\left(|y_i|,85-x_i\right)\frac{180}{\pi}.
\label{eq:shot_angle}
\end{equation}

Distance is measured in feet and angle in degrees; smaller angles lie closer to the center line. Angles above $90^\circ$ identify the small number of attempts originating behind the goal line, which were also marked by a separate indicator. Raw coordinates and the inferred attacked-goal coordinate were retained for audit but excluded from the model.

\subsection{Temporal Features}

Temporal variables consisted of period, time remaining in seconds, shot-clock seconds, and an indicator for a disabled shot clock. Separating the indicator prevented the raw $-1$ sentinel from being interpreted as possession time. These variables were intended to capture changes in offensive urgency and shot selection over a possession or game.

\subsection{Situational Context Features}

Situational context comprised strength state, offense type, score differential, and pulled-goalie state. Strength identified even-strength, power-play, short-handed, extra-attacker, and penalty-shot situations. Offense type distinguished transition from settled possessions. A passer-present indicator represented whether a final passer was recorded without conditioning the probability on passer identity.

\subsection{Offensive Context Features}

Offensive-context variables described how the shot developed. Shot type recorded overhand, underhand, sidearm, quick-stick, shovel, dunk, behind-the-back, twister, or wraparound attempts. Validated swing status was consolidated into \textit{off swing} and \textit{not off swing}; off-pass status distinguished \textit{shot directly off pass} from \textit{self-created}.

The two-man-action variable recorded the shooter's role as \textit{ball carrier}, \textit{picker}, or \textit{not two man action}. Qualifying pick type recorded \textit{on ball}, \textit{off ball}, or \textit{no qualifying pick}. These two fields formed the intervention used for counterfactual xPV. Picker number identified the player receiving attribution but was never a model feature.

\subsection{Outcome Variable}

The four recorded outcomes were converted to the binary target

\begin{equation}
y_i =
\begin{cases}
1, & \text{if shot $i$ resulted in a goal},\\
0, & \text{if shot $i$ was saved, missed, or blocked}.
\end{cases}
\label{eq:binary_outcome}
\end{equation}

The models estimated the probability that this target equaled one. The three nested feature specifications tested whether the two action fields improved held-out probability quality beyond the contextual baseline.

\section{Machine Learning Methodology}
\label{sec:methodology}

\subsection{Problem Formulation}

For shot features $\mathbf{x}_i$ and binary outcome $y_i$ (Equation~\ref{eq:binary_outcome}), the expected-goals model estimates

\begin{equation}
\mathrm{xG}_i=\Pr\left(y_i=1\mid\mathbf{x}_i\right).
\label{eq:xg_definition}
\end{equation}

This probability is the xG assigned to the shot and describes opportunity quality independently of the observed outcome.

\subsection{Data Preprocessing}

\subsubsection{Feature Engineering}

An automated pipeline generated the variables in Section~\ref{sec:event_representation} using deterministic transformations. It normalized attacking direction by game and period; calculated distance, angle, time, sentinel indicators, and categorical groupings; and validated the relationships among picker number, pick type, and two-man-action state. Raw attacking-goal metadata were used only for spatial transformation and audit.

\begin{figure}[t]
    \centering
    \includegraphics[width=\linewidth]
        {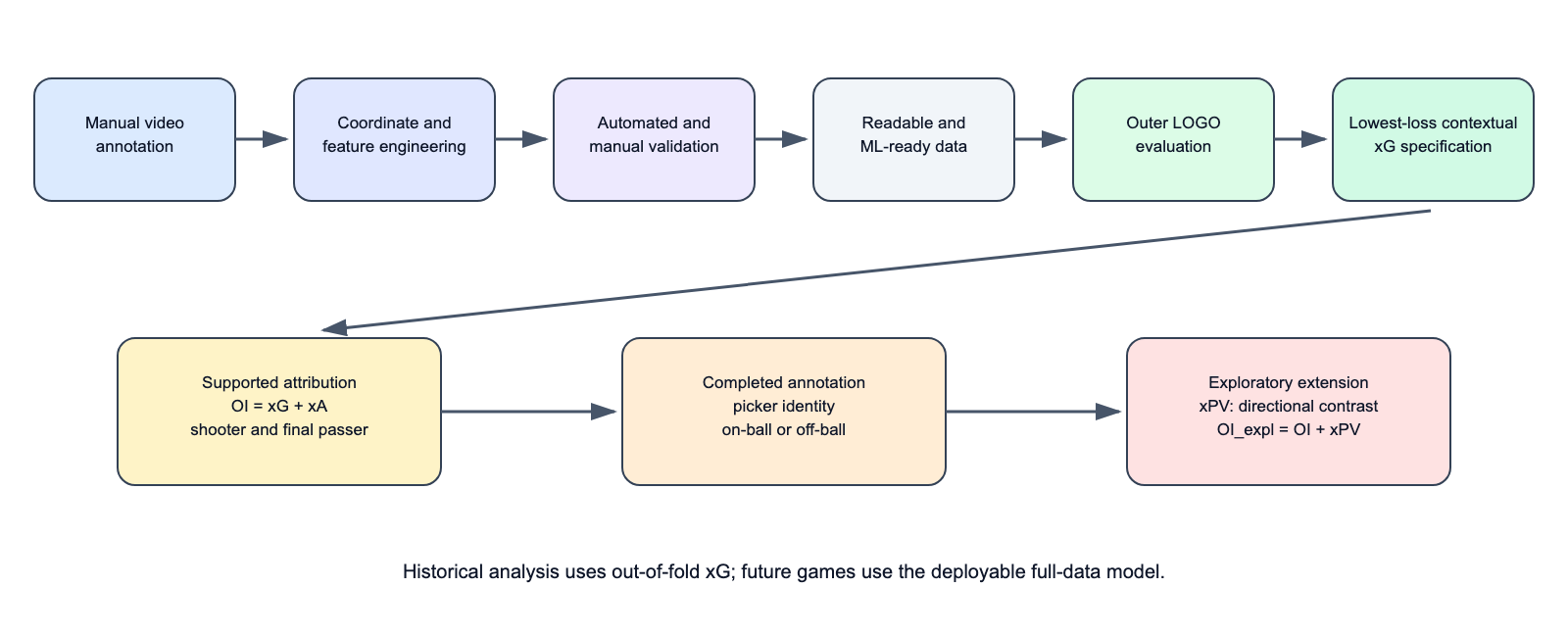}
    \caption{Data-preparation and modeling workflow. Manually annotated events
    undergo feature engineering, validation, and encoding before model
    development and downstream offensive analysis.}
    \label{fig:feature_pipeline}
\end{figure}

\subsubsection{Feature Validation}

Attacking direction was checked within every game-period group, representative coordinate transformations were inspected, and distance and angle ranges were tested for geometric consistency. Categorical mappings were reviewed, while automated checks compared raw and processed record counts and identified missing values, invalid ranges, unexpected categories, and inconsistent sentinel combinations. All 1,006 raw records produced validated processed observations.

\subsubsection{Feature Encoding}

Within each training fold, missing numeric values were median-imputed and standardized; categorical values were most-frequent imputed and one-hot encoded with unseen held-out levels ignored. Thus a context left unassigned during video review was handled by the prespecified fold-level imputation rule rather than treated as an observed category. Because visibility-related missingness may not be random, this choice is a limitation of the present analysis. Binary indicators were retained numerically. Identifiers, outcomes, raw horizontal coordinate, attacking-goal metadata, and attribution-only picker number were excluded from the predictor matrix.

\subsection{Machine Learning Models Evaluated}

Three fixed classifiers were compared using predicted probabilities under the same validation design. Logistic regression \citep{hosmer2013applied} provided a regularized linear benchmark. Random forest \citep{breiman2001random} averaged bootstrapped decision trees, while extremely randomized trees added greater split randomization \citep{geurts2006extremely}. Both ensembles could represent nonlinear relationships and interactions. The comparison was designed to evaluate the specified pipelines transparently rather than to exhaustively tune every algorithm.

Each classifier was crossed with three nested feature sets. The contextual baseline contained the ten numeric and seven categorical source fields listed in Table~\ref{tab:event_features}. The second set added two-man-action role. The third added qualifying pick type and was the only specification capable of expressing the complete counterfactual xPV intervention.

\subsection{Cross-Validation Strategy}
\label{subsec:crossvalidationstrategy}

Performance was evaluated using Leave-One-Game-Out (LOGO) cross-validation because shots within a game share opponents, goaltenders, score progression, and tactical conditions. A random shot-level split could place correlated observations from the same game in both training and testing data.

For outer fold $i$,

\begin{equation}
\begin{aligned}
\mathcal{D}_{\mathrm{train}}^{(i)}
&=\bigcup_{\substack{j=1 \\ j\neq i}}^{13}\mathcal{D}_{g_j}, &
\mathcal{D}_{\mathrm{test}}^{(i)}&=\mathcal{D}_{g_i}.
\end{aligned}
\label{eq:logo_splits}
\end{equation}

Each game served as the test set once. All imputation, encoding, scaling, and model fitting were restricted to the other twelve games. The prespecified fixed pipelines were fitted directly within each training fold and evaluated on the isolated game. Concatenating the thirteen test sets gave every historical shot an out-of-fold probability from a model that had seen no observations from its game.

The same aggregate out-of-fold predictions were then used to compare the nine fixed specifications and identify the lowest-loss specification. Consequently, Table~\ref{tab:full_model_comparison} is a descriptive cross-validated comparison, not an independent or nested post-selection performance estimate. The modest differences should not be interpreted as statistically established gaps.

\subsection{Prespecified Model Settings}

The experiment used prespecified settings rather than nested optimization. Logistic regression used $C=1$, no class weighting, and a 3,000-iteration limit. Random forest and extremely randomized trees each used 500 trees, minimum terminal-node size five, and the square root of the encoded feature count at each split. Neither ensemble used class weighting. All stochastic estimators used seed 20251214. Complete settings appear in Appendix~\ref{app:hyperparameters}.

\subsection{Evaluation Metrics}

Metrics were calculated from the combined out-of-fold predictions. Let $p_i$ denote the probability assigned to shot $i$, $y_i$ its outcome, and $N$ the number of shots.

\subsubsection{Log Loss}

Log loss, the primary selection metric, was

\begin{equation}
\mathrm{LogLoss}=-\frac{1}{N}\sum_{i=1}^{N}
\left[y_i\log(p_i)+(1-y_i)\log(1-p_i)\right].
\label{eq:log_loss}
\end{equation}

Lower values indicate better probabilities and confident errors receive larger penalties.

\subsubsection{Brier Score}

The secondary probability metric was the Brier score \citep{brier1950verification},

\begin{equation}
\mathrm{Brier}=\frac{1}{N}\sum_{i=1}^{N}(p_i-y_i)^2,
\label{eq:brier_score}
\end{equation}

for which lower values are better.

\subsubsection{Receiver Operating Characteristic Area Under the Curve}

ROC AUC measured the probability that a goal was ranked above a non-goal \citep{Hanley1982}. Values of 0.50 and 1.00 indicate chance and perfect discrimination, respectively. Because ROC AUC does not assess numerical probability accuracy, it was supporting rather than primary evidence. It was not interpreted for the fold-specific constant benchmark.

\subsubsection{Average Precision}

Average precision summarized the precision-recall relationship and was included because goals comprised approximately 15\% of observations \citep{davis2006relationship}. It measures positive-class ranking, not probability accuracy, and was interpreted with log loss and Brier score.

\subsubsection{Calibration}

Calibration used ten equal-width probability intervals \citep{niculescu2005predicting,guo2017calibration}. Expected calibration error was

\begin{equation}
\mathrm{ECE}=\sum_{b=1}^{B}\frac{n_b}{N}
\left|\overline{y}_b-\overline{p}_b\right|,
\label{eq:ece}
\end{equation}

where $n_b$, $\overline{y}_b$, and $\overline{p}_b$ are the size, observed rate, and mean prediction in interval $b$. Because bin-based ECE depends on the distribution of predictions, it was treated as supporting evidence rather than a replacement for log loss or Brier score.

\subsection{Final Model Selection}

The nine algorithm--feature-set combinations were ranked descriptively by aggregate out-of-fold log loss, followed by Brier score and ten-bin ECE as tie-breakers. Section~\ref{sec:model_results} identifies the contextual baseline random forest as the lowest-loss general xG specification among those evaluated. Counterfactual xPV required both intervention variables, so the lowest-loss full pick-aware specification was retained separately for that exploratory task. Both deployable pipelines were refitted on all 1,006 observations, while historical attribution used out-of-fold probabilities.

\section{Model Results}
\label{sec:model_results}

Results were calculated from LOGO out-of-fold probabilities for 1,006 shots (151 goals; observed rate 15.01\%).

\subsection{Base-Rate Benchmark}
\label{subsec:base_rate_benchmark}

Because goals are infrequent, a model can achieve superficially low log loss and Brier score without using any shot information. All specifications were therefore compared against a constant predictor that assigns every held-out shot the goal rate of its training fold. This benchmark uses the identical LOGO partition and no features. Pooled across folds it achieved log loss 0.4241 and Brier score 0.1279.

The benchmark is reported alongside every specification in Table~\ref{tab:full_model_comparison} and should govern interpretation of the comparison. Differences among the nine specifications are small relative to the distance between the lowest-loss specification and the benchmark, and that distance is itself modest.

\subsection{Nine-Model Comparison}
\label{subsec:full_model_comparison}

The contextual baseline random forest had the lowest observed aggregate out-of-fold log loss and Brier score (Table~\ref{tab:full_model_comparison}). All three random-forest specifications ranked ahead of the extremely randomized trees and logistic-regression pipelines. The comparison covers the specified fixed configurations and does not establish that random forest will dominate under every tuning strategy or larger dataset.

The lowest-loss specification improved on the base-rate benchmark by 1.22\% in log loss and 1.50\% in Brier score. Four of the nine specifications improved on the benchmark and five did not: both extremely randomized trees extensions and all three logistic-regression pipelines produced worse held-out log loss than predicting the training-fold goal rate. Because these same predictions governed model selection, the percentages are descriptive and are not independent post-selection performance estimates. The recorded shot context therefore supplies only limited probability information beyond the base rate in this sample. This is consistent with the ROC AUC of 0.604, which indicates weak separation between goals and non-goals.

Metrics are reported to four decimal places. The differences among the three random-forest specifications occur in the fourth decimal and are smaller than the game-level variation documented in Section~\ref{subsec:game_level_uncertainty}; they should not be read as established performance gaps.

\begin{table}[ht]
\centering
\small
\setlength{\tabcolsep}{4pt}
\caption{Complete out-of-fold comparison against a base-rate benchmark. Lower log loss, Brier score, and ECE are better; higher ROC AUC and average precision (AP) are better. $\Delta$LL is the percentage reduction in log loss relative to the benchmark, with negative values indicating performance worse than predicting the training-fold goal rate.}
\label{tab:full_model_comparison}
\begin{tabular}{llrrrrrr}
\toprule
Feature set & Model & Log Loss & $\Delta$LL & Brier & ROC AUC & AP & ECE \\
\midrule
Baseline & Random forest & 0.4189 & $+1.22\%$ & 0.1260 & 0.604 & 0.214 & 0.026 \\
$+$ two-man action & Random forest & 0.4195 & $+1.07\%$ & 0.1261 & 0.603 & 0.215 & 0.019 \\
$+$ pick type & Random forest & 0.4197 & $+1.03\%$ & 0.1260 & 0.604 & 0.216 & 0.019 \\
Baseline & Extra Trees & 0.4219 & $+0.51\%$ & 0.1268 & 0.569 & 0.207 & 0.030 \\
$+$ two-man action & Extra Trees & 0.4256 & $-0.37\%$ & 0.1277 & 0.557 & 0.195 & 0.027 \\
$+$ pick type & Extra Trees & 0.4269 & $-0.66\%$ & 0.1276 & 0.553 & 0.204 & 0.035 \\
Baseline & Logistic regression & 0.4311 & $-1.65\%$ & 0.1299 & 0.582 & 0.205 & 0.050 \\
$+$ pick type & Logistic regression & 0.4315 & $-1.76\%$ & 0.1299 & 0.584 & 0.211 & 0.049 \\
$+$ two-man action & Logistic regression & 0.4328 & $-2.07\%$ & 0.1303 & 0.579 & 0.203 & 0.050 \\
\midrule
None & Base-rate benchmark & 0.4241 & --- & 0.1279 & --- & --- & --- \\
\bottomrule
\end{tabular}
\end{table}

\subsection{Effect of Pick Features}
\label{subsec:pick_feature_extension}

Within random forest, adding two-man-action role increased log loss from 0.4189 to 0.4195 and increased Brier score from 0.1260 to 0.1261. Adding qualifying pick type increased log loss further to 0.4197, while Brier score remained slightly above baseline. ROC AUC and average precision changed only modestly. ECE was lower for both extensions, but the improvement in this bin-dependent supporting metric did not offset the worse primary probability metrics. The contextual baseline was therefore selected for general xG prediction.

These results do not imply that picks are unrelated to shot creation. They show only that the two recorded action fields did not improve aggregate out-of-game probability quality in the present sample after the remaining shot context was included. Limited statistical information is one plausible mechanism: the one-hot encoded predictors are sparse relative to the 151 observed goals, and only fifteen off-ball picks were available. That category in particular cannot support reliable estimation.

\subsection{Validation of Counterfactual xPV}
\label{subsec:xpv_validation}

Because the pick fields did not improve held-out probability quality, a pick-aware model may still produce nonzero counterfactual contrasts purely through arbitrary partitioning on those fields. Two diagnostics were therefore run to establish whether the reported xPV values can be distinguished from model noise.

The first is a placebo intervention. For the 781 shots recorded without a qualifying pick, the two pick fields were set \emph{to} the most common pick state and the contrast recomputed. These shots contain no pick, so any movement is definitionally noise. The mean absolute placebo contrast was 0.0176, compared with 0.0222 for the 225 genuine qualifying picks.

The second is a row-permutation diagnostic. Within each training fold the two pick fields were permuted jointly across rows, destroying their row-level relationship with the outcome and the remaining context while preserving their marginal distribution and their association with each other. The pipeline was refit for every fold and replicate, and the intervention was applied to the same held-out qualifying-pick shots used for observed xPV. Two hundred replicates were generated using the analysis seed. This permutation was performed across rows rather than within games, so it does not preserve game-level pick composition; together with the limited tail resolution of 200 replicates, that design makes the exercise an exploratory model diagnostic rather than a confirmatory randomization test. Table~\ref{tab:xpv_validation} therefore reports comparison counts rather than formal $p$ values.

\begin{table}[ht]
\centering
\small
\caption{Exploratory counterfactual xPV diagnostic in which the pick fields are permuted jointly across rows within each training fold. Two hundred replicates, seed 20251214. The permutation does not preserve game-level pick composition; comparison counts are descriptive rather than confirmatory $p$ values.}
\label{tab:xpv_validation}
\begin{tabular}{lrrrl}
\toprule
Statistic & Observed & Null mean & Null 95\% interval & Diagnostic count \\
\midrule
Mean $|$xPV$|$ & 0.0222 & 0.0236 & $[0.0181,\ 0.0306]$ & 74/200 below observed \\
Fraction positive & 0.822 & 0.551 & $[0.391,\ 0.707]$ & 0/200 at or above observed \\
\bottomrule
\end{tabular}
\end{table}

The two diagnostics separate two aspects of the measure. The \emph{magnitude} of xPV is indistinguishable from model noise: the mean absolute contrast under the permutation diagnostic was slightly larger than the observed value, 74 of the 200 replicates produced a smaller mean absolute contrast than the real data, and the placebo intervention on no-pick shots produced contrasts of comparable size. The placebo also moved positive for 76.6\% of those shots, so the model's directional response to the pick state is not specific to shots where a pick was actually recorded. Individual xPV values and player xPV totals therefore cannot be interpreted as calibrated goal-equivalent quantities.

The \emph{direction} is more consistent than this particular shuffled-feature benchmark. Positive contrasts occurred for 82.2\% of qualifying picks against a replicate mean of 55.1\%, and none of the 200 replicates reached the observed value; the largest was 76.9\%. The shuffled benchmark is itself directionally biased, so the relevant descriptive comparison is with that distribution rather than with an even split. Every one of the thirteen held-out games produced positive contrasts for more than half its qualifying picks, ranging from 54.5\% to 100\%. These findings show how the fitted pick-aware model responds to the recorded fields; they do not establish a causal pick effect or a cluster-valid inferential probability. A future league-wide analysis should repeat the diagnostic with game-preserving permutations and enough replicates for stable tail estimation.

A further limitation constrains what player-level xPV can express. Because picker identity is never a model feature and the pick state is represented by only two categorical fields, xPV for a given attempt is determined by the recorded shot context rather than by who set the pick. Across the nine pickers credited with at least five qualifying picks, mean xPV per pick ranged only from 0.0156 to 0.0301, and player xPV totals correlated with raw pick count at Spearman $\rho = 0.98$. An additional picker-label permutation exercise produced between-player variation in mean per-pick xPV that was no smaller than the observed variation (0.0063 under permutation against 0.0050 observed). Player xPV totals are accordingly close to a rescaled pick count and should not be read as evidence that one player's picks generate more value than another's. Distinguishing pickers would require annotating pick execution itself, such as defensive coverage response, contact, and separation created.

Two further caveats apply. The permutation breaks the association between the pick fields and the remaining shot context, while its across-row implementation also ignores game blocks. Its direction of bias is therefore not assumed. The contrast also holds shot location fixed, so it cannot capture value operating through the pick producing a better location in the first place. Counterfactual xPV is accordingly reported throughout as an exploratory directional contrast, and player xPV totals should not be compared as if they were measured quantities.

\subsection{Game-Level Variation}
\label{subsec:game_level_uncertainty}

Baseline random-forest log loss varied from 0.295 to 0.597 across the thirteen held-out games. The baseline achieved lower game-level log loss than the full pick-aware random forest in eight of thirteen games. This variation exceeds the aggregate differences among the three random-forest specifications and reinforces the need for additional games and independent teams before small aggregate differences are treated as stable.

\section{Offensive Impact Framework}
\label{sec:offensive_impact}

\subsection{Framework Overview}
\label{subsec:oi_overview}

The Offensive Impact (OI) framework uses shot-level xG as a common goal-equivalent unit for attributing supported recorded offensive involvement through two roles: the shooter and the recorded final passer. A third role, the qualifying picker, is represented separately by exploratory counterfactual xPV. Finishing Above Expected (FAE) is also reported separately so that opportunity involvement is not conflated with the outcome of a player's shots.

The attribution is intentionally shared. One attempt may generate xG for the shooter and xA for the final passer, while an exploratory xPV contrast may be associated with a qualifying picker. These values describe different recorded roles in the same opportunity and are not independent additions to team expected scoring. Team opportunity value remains the sum of shot-level xG.

Historical xG and xA use Leave-One-Game-Out out-of-fold probabilities. The completed annotations identify shooters, final passers, qualifying non-shooting pickers, and on-ball or off-ball pick type. The empirical attribution extension uses the observed-state out-of-fold probability from the pick-aware model for both shooter xG and passer xA, keeping those quantities on the same probability scale as the counterfactual xPV contrast. Picker identity is retained only for attribution and is never a model predictor.

\subsection{Expected-Goal Generation}
\label{subsec:expected_goal_generation}

For player $p$, total expected-goal generation is

\begin{equation}
\mathrm{xG}_p
=
\sum_{i\in S_p}\mathrm{xG}_i,
\label{eq:player_xg}
\end{equation}

where $S_p$ is the set of shots taken by player $p$. If $N_p=|S_p|$ is the number of attempts, average xG per shot is

\begin{equation}
\overline{\mathrm{xG}}_p
=
\frac{\mathrm{xG}_p}{N_p}.
\label{eq:average_xg_per_shot}
\end{equation}

Total xG reflects both opportunity volume and modeled opportunity quality, whereas average xG per shot describes mean opportunity quality. Both should be reported with shot count. A player's share of team shot value is

\begin{equation}
\mathrm{xGShare}_p
=
\frac{\mathrm{xG}_p}{\sum_{q\in\mathcal{R}}\mathrm{xG}_q},
\label{eq:player_xg_share}
\end{equation}

where $\mathcal{R}$ is the set of represented rostered players. Shot-based xG can also be separated into self-created attempts and shots taken directly from passes. Crediting xG to the shooter identifies who completed the attempt; it does not imply that the shooter created the opportunity alone. Shooter identity was not a default predictor, so the model estimates the recorded opportunity without directly conditioning on the shooter's name or jersey number.

\subsection{Finishing Above Expected}
\label{subsec:finishing_above_expected}

FAE compares observed goals with the modeled value of a player's shots:

\begin{equation}
\mathrm{FAE}_p
=
G_p-\mathrm{xG}_p,
\label{eq:finishing_above_expected}
\end{equation}

where $G_p$ is the observed goal total. Positive FAE indicates more goals than expected and negative FAE indicates fewer. A per-shot form is

\begin{equation}
\mathrm{FAEPerShot}_p
=
\frac{G_p-\mathrm{xG}_p}{N_p},
\label{eq:fae_per_shot}
\end{equation}

and the ratio-based Finishing Index is

\begin{equation}
\mathrm{FI}_p
=
\frac{G_p}{\mathrm{xG}_p}.
\label{eq:finishing_index}
\end{equation}

A Finishing Index of 1.00 indicates scoring at expectation. Because ratios are unstable when xG is small, each finishing measure should be accompanied by goals, xG, and shot count. FAE describes observed conversion within the analyzed sample; it should not be treated as persistent shooting ability without substantially more attempts.

At team level,

\begin{equation}
\mathrm{FAE}_{\mathrm{team}}
=
G_{\mathrm{team}}-\mathrm{xG}_{\mathrm{team}},
\label{eq:team_fae}
\end{equation}

which helps distinguish limited opportunity generation from conversion above or below the modeled expectation.

\subsection{Expected Assists and Opportunity Creation}
\label{subsec:expected_assists}

Expected assists measure the expected scoring value associated with a player's recorded final passes. For player $p$,

\begin{equation}
\mathrm{xA}_p
=
\sum_{i\in A_p}\mathrm{xG}_i,
\label{eq:player_xa}
\end{equation}

where $A_p$ is the set of shots for which player $p$ was identified as the final passer. A pass generates xA whether the shot becomes a goal, save, miss, or block; attempts without an identified passer generate no xA. This separates final-pass opportunity creation from the shooter's finishing.

If $N_p^{\mathrm{SA}}=|A_p|$ is the number of shot assists, average xA per shot assist is

\begin{equation}
\overline{\mathrm{xA}}_p
=
\frac{\mathrm{xA}_p}{N_p^{\mathrm{SA}}}.
\label{eq:average_xa_per_shot_assist}
\end{equation}

Total xA reflects the frequency and quality of shot-producing passes; the average describes the mean modeled quality of the resulting attempts. For a passer--shooter pair, $A_{p\rightarrow q}$ denotes shots taken by player $q$ after a final pass from player $p$, giving

\begin{equation}
\mathrm{xA}_{p\rightarrow q}
=
\sum_{i\in A_{p\rightarrow q}}\mathrm{xG}_i.
\label{eq:pairwise_xa}
\end{equation}

Pairwise xA should be reported with linked-shot count. The measure credits only the final passer and excludes earlier passes and passes that do not produce shots; it is therefore a shot-creation measure rather than overall passing efficiency.

\subsection{Per-Event Rates and Finishing}
\label{subsec:offensive_efficiency}

Cumulative values are influenced by opportunity volume. Observed shooting percentage is

\begin{equation}
\mathrm{ShootingPercentage}_p
=
\frac{G_p}{N_p},
\label{eq:shooting_percentage}
\end{equation}

while average xG per shot describes average modeled opportunity quality. Their difference equals FAE per shot:

\begin{equation}
\mathrm{FAEPerShot}_p
=
\mathrm{ShootingPercentage}_p-\overline{\mathrm{xG}}_p.
\label{eq:fae_per_shot_decomposition}
\end{equation}

Average xA per shot assist provides the corresponding rate for final-pass creation. Average xPV per qualifying pick describes the mean counterfactual probability contrast across credited actions. Totals, rates, and event counts should be presented together because high per-event values based on few involvements can be misleading. Complete possession, shift, and time-on-floor data were unavailable, preventing per-possession or per-minute adjustment.

\subsection{Definition of Offensive Impact}
\label{subsec:offensive_impact_definition}

Core Offensive Impact combines the supported expected value attributed through shooting and final passing:

\begin{equation}
\mathrm{OI}_p
=
\mathrm{xG}_p+\mathrm{xA}_p.
\label{eq:offensive_impact}
\end{equation}

FAE is not included in Equation~\ref{eq:offensive_impact}; it remains a separate measure of conversion. This preserves OI as an opportunity-involvement measure and prevents short-term finishing from obscuring how expected value was generated.

Shooter xG and passer xA are out-of-fold predicted probabilities on a common scale. For descriptive illustration only, an exploratory augmented score is

\begin{equation}
\mathrm{OI}^{\mathrm{expl}}_p
=
\mathrm{OI}_p+\mathrm{xPV}_p.
\label{eq:exploratory_offensive_impact}
\end{equation}

The xPV term is a counterfactual model contrast whose magnitude, as Section~\ref{subsec:xpv_validation} reports, cannot be distinguished from model noise. Equation~\ref{eq:exploratory_offensive_impact} is retained to show how picker credit could enter a better-resolved future framework, not as a validated goal-equivalent player score. Empirical rankings based on this exploratory augmented score are labeled accordingly.

\subsubsection{Expected Pick Value}

Expected Pick Value estimates the change in predicted scoring probability associated with the recorded pick state while holding the remaining observed shot record fixed. An on-ball pick qualifies when the ball carrier uses the screen and shoots within the same continuous action; an off-ball pick qualifies when a teammate uses the screen, receives a pass, and shoots within the same continuous action.

For qualifying shot $i$, let $\widehat p_i^{\mathrm{obs}}$ be the pick-aware model's out-of-fold probability for the recorded row. A counterfactual copy is created by changing only the two-man-action role to \texttt{not two man action} and the qualifying pick type to \texttt{no\_qualifying\_pick}. Shot location, shooter, passer, time, strength, score state, and all remaining recorded context are retained. If $\widehat p_i^{\mathrm{no\mbox{-}pick}}$ denotes the probability assigned to the modified row, then

\begin{equation}
\mathrm{xPV}_i
=
\widehat p_i^{\mathrm{obs}}
-
\widehat p_i^{\mathrm{no\mbox{-}pick}}.
\label{eq:shot_level_xpv}
\end{equation}

Non-qualifying shots and rows without a valid picker receive zero xPV. For the set $P_p^{\mathrm{on}}$ of qualifying on-ball attempts credited to picker $p$,

\begin{equation}
\mathrm{xPV}_p^{\mathrm{on}}
=
\sum_{i\in P_p^{\mathrm{on}}}\mathrm{xPV}_i.
\label{eq:on_ball_xpv}
\end{equation}

For the corresponding off-ball set $P_p^{\mathrm{off}}$,

\begin{equation}
\mathrm{xPV}_p^{\mathrm{off}}
=
\sum_{i\in P_p^{\mathrm{off}}}\mathrm{xPV}_i,
\label{eq:off_ball_xpv}
\end{equation}

and total xPV is

\begin{equation}
\mathrm{xPV}_p
=
\mathrm{xPV}_p^{\mathrm{on}}+\mathrm{xPV}_p^{\mathrm{off}}.
\label{eq:total_xpv}
\end{equation}

Unused picks, actions followed by an offensive reset, ambiguous sequences, and pick-and-roll or pick-and-pop sequences in which the picker becomes the shooter receive no picker credit. If multiple picks occur, only the final qualifying non-shooting picker directly involved in the attempt is credited.

This contrast is not a causal treatment effect. Shot location and the remaining observed context may themselves be consequences of the pick, so holding them fixed estimates a conditional direct contrast rather than the pick's total possession-level value. Unmeasured pressure, defensive rotation, and selection into pick actions may also affect both the recorded action and the resulting shot. The model comparison in Section~\ref{subsec:pick_feature_extension} is therefore essential context for interpreting empirical xPV.

\subsection{Aggregation and Interpretation}
\label{subsec:oi_aggregation}

Game-level core OI for player $p$ in game $g$ is

\begin{equation}
\mathrm{OI}_{p,g}
=
\sum_{i\in S_{p,g}}\mathrm{xG}_i
+
\sum_{i\in A_{p,g}}\mathrm{xG}_i,
\label{eq:game_level_oi}
\end{equation}

where the two sets contain the player's shots and shot-producing final passes. Exploratory game-level augmented OI adds $\sum_{i\in P_{p,g}}\mathrm{xPV}_i$, where $P_{p,g}$ contains qualifying picks. A higher core OI indicates more expected scoring value associated with recorded shooting and passing involvement, not necessarily greater per-event quality. Similar totals may arise through different roles, so each component and its event count should remain visible.

OI is not conserved across players. Summing individual values counts team shot-based xG once and then adds shared attribution to final passers; exploratory augmented OI additionally includes picker contrasts. Team scoring performance should therefore be evaluated using goals, shot-based xG, and team FAE. OI describes how recorded involvement in those opportunities is distributed among players and should not be interpreted as unique marginal contribution.

\section{Empirical Offensive Impact}
\label{sec:empirical_oi}

\subsection{Player Offensive Impact}

Table~\ref{tab:player_oi_results} reports the nine Rochester players credited with at least five qualifying picks, which isolates the regular offensive rotation: every remaining player in the attribution dataset was credited with at most one. The displayed ranks use the exploratory augmented score in Equation~\ref{eq:exploratory_offensive_impact}; they are descriptive and should not be interpreted as validated player-value rankings. The complete 25-player attribution roster appears in Appendix~\ref{app:full_attribution}. Names and jersey identifiers were reconciled against the official Rochester Knighthawks roster \citep{rochesterroster2026}. Total xG measures cumulative expected scoring contribution from a player's shots, whereas xG per shot measures the average modeled quality of those opportunities. It is not a measure of finishing efficiency, which requires comparing actual goals with expected goals. The criterion is a proxy for membership in the settled-offence rotation rather than a threshold chosen for the shooting comparisons that follow; the same nine players are the sample's primary shooters, and Appendix~\ref{app:full_attribution} gives every player for readers who prefer an unfiltered view.

\begin{table}[ht]
\centering
\small
\setlength{\tabcolsep}{4pt}
\caption{Exploratory augmented Offensive Impact for the nine Rochester players credited with at least five qualifying picks. Rank is calculated among all 25 players in the attribution dataset (Appendix~\ref{app:full_attribution}). xG/shot is observed-state out-of-fold xG divided by attempts; $\mathrm{OI}^{\mathrm{expl}}$ equals xG $+$ xA $+$ xPV. Because xPV magnitudes are not distinguishable from model noise, neither the xPV column nor the augmented ranking is a validated goal-equivalent player comparison.}
\label{tab:player_oi_results}
\begin{tabular}{@{}lrrrrrrr@{}}
\toprule
\textbf{Player (No.)} & \textbf{Rank} & \textbf{Shots} & \textbf{xG/shot} & \textbf{xG} & \textbf{xA} & \textbf{xPV} & $\mathbf{OI^{expl}}$ \\
\midrule
Ryan Lanchbury (7) & 1 & 131 & 0.144 & 18.83 & 45.27 & 0.21 & 64.32 \\
Connor Fields (10) & 2 & 233 & 0.140 & 32.56 & 27.59 & 0.15 & 60.30 \\
Ryan Smith (43) & 3 & 194 & 0.160 & 31.12 & 12.46 & 0.76 & 44.34 \\
Thomas McConvey (12) & 4 & 165 & 0.161 & 26.57 & 10.91 & 1.16 & 38.64 \\
Zed Williams (28) & 5 & 93 & 0.156 & 14.47 & 6.58 & 0.71 & 21.76 \\
Kyle Waters (42) & 6 & 43 & 0.144 & 6.18 & 4.39 & 0.62 & 11.20 \\
Graydon Hogg (16) & 7 & 45 & 0.141 & 6.35 & 3.69 & 0.27 & 10.31 \\
Brad McCulley (20) & 13 & 15 & 0.166 & 2.49 & 0.90 & 0.27 & 3.66 \\
Blaze Riorden (11) & 14 & 9 & 0.130 & 1.17 & 1.52 & 0.11 & 2.80 \\
\bottomrule
\end{tabular}
\end{table}

Table~\ref{tab:player_counting_results} provides the corresponding conventional production counts. ``Assists'' refers to goals for which the player was identified as the recorded final passer in the annotated data; it does not include a possible secondary assist under official scoring rules.

\begin{table}[ht]
\centering
\small
\caption{Shots, goals, and recorded final-pass assists for the same nine players.}
\label{tab:player_counting_results}
\begin{tabular}{@{}lrrrr@{}}
\toprule
\textbf{Player (No.)} & \textbf{Expl. rank} & \textbf{Shots} & \textbf{Goals} & \textbf{Assists} \\
\midrule
Ryan Lanchbury (7) & 1 & 131 & 22 & 40 \\
Connor Fields (10) & 2 & 233 & 31 & 24 \\
Ryan Smith (43) & 3 & 194 & 27 & 17 \\
Thomas McConvey (12) & 4 & 165 & 24 & 12 \\
Zed Williams (28) & 5 & 93 & 18 & 4 \\
Kyle Waters (42) & 6 & 43 & 4 & 6 \\
Graydon Hogg (16) & 7 & 45 & 6 & 5 \\
Brad McCulley (20) & 13 & 15 & 0 & 0 \\
Blaze Riorden (11) & 14 & 9 & 3 & 3 \\
\bottomrule
\end{tabular}
\end{table}

Ryan Lanchbury (No. 7) ranked first on the exploratory augmented score, with 45.27 xA identifying a high-volume final-pass creation role. Connor Fields (No. 10) ranked second and generated the largest total xG at 32.56. Ryan Smith (No. 43) followed with 31.12 xG. Thomas McConvey (No. 12) had the largest descriptive xPV total at 1.16, but that value should not be interpreted as evidence that his picks were more effective than another player's. The four highest-ranked players accounted for 71.0\% of the displayed augmented-score sum, describing a concentration of recorded involvement rather than unique marginal impact.

Fields's leading total xG was driven primarily by shooting volume. He recorded 233 attempts, substantially more than Smith (194), McConvey (165), Lanchbury (131), or Zed Williams (93). His xG per shot was 0.140, the lowest among these five high-volume shooters. Fields therefore accumulated substantial expected scoring value through frequent shooting involvement rather than through an unusually valuable average attempt. This distinction demonstrates why total xG and xG per shot should be interpreted together: total xG measures accumulated opportunity value, whereas xG per shot measures the quality of the average opportunity.

Ranking the five high-volume shooters by xG per shot produces a different ordering. McConvey ranked first at 0.161, followed by Smith at 0.160, Williams at 0.156, Lanchbury at 0.144, and Fields at 0.140. McConvey and Smith therefore generated the highest average-quality opportunities among the primary shooters, while Williams also produced a comparatively high-quality average shot profile. A difference of 0.010 xG per shot corresponds to approximately one additional expected goal per 100 attempts, holding shooting volume constant.

Williams provides a particularly useful example of the difference between cumulative and per-shot value. His 0.156 xG per shot was higher than the rates recorded by Lanchbury and Fields and only slightly lower than those of McConvey and Smith. His total xG was nevertheless lower at 14.47 because he attempted only 93 shots. Williams therefore generated relatively strong expected value when he shot but accumulated less total xG because of lower shooting volume. His 18 goals also exceeded his 14.47 xG by approximately 3.53 goals during the analyzed games. That finishing result is conceptually separate from his xG-per-shot ranking: xG per shot describes the modeled quality of his average opportunity, whereas goals above expected describes how successfully he converted those opportunities.

Across all nine selected offensive players, McCulley recorded the highest xG per shot at 0.166, followed by McConvey (0.161), Smith (0.160), Williams (0.156), Waters (0.144), Lanchbury (0.144), Hogg (0.141), Fields (0.140), and Riorden (0.130). McCulley's rate, however, was based on only 15 attempts, while Riorden's was based on nine. These small samples are less stable than the rates for the primary shooters and should not be treated as established differences in player skill.

More generally, xG per shot reflects shot location and the context represented by the model rather than whether the player converted the chance. A high rate may reflect strong shot selection, effective off-ball positioning, or an offensive role that produces close-range opportunities. It does not by itself establish superior shooting talent or finishing ability. For that reason, xG per shot should always be presented alongside total xG, shot count, goals, and finishing relative to expectation.

\section{Practical Applications}
\label{sec:applications}

The framework produces xG, xA, core OI, and exploratory counterfactual xPV outputs from any annotated game, supporting three applications. Each is constrained by the probability quality documented in Section~\ref{sec:model_results}: because the lowest-loss evaluated model improves on the base rate by roughly one percent, these outputs organize evidence for film review rather than replace it.

\paragraph{Player and game profiles.}
A profile presents volume, attributed expected value, and per-event rates together: attempts, goals, xG, average xG per shot, shooting percentage, FAE, shot assists, xA, core OI, qualifying picks, and exploratory xPV. Keeping the components separate prevents a combined score from concealing role. FAE remains separate because it describes conversion rather than opportunity involvement, and every rate should be accompanied by its event count. At game level the framework also gives a fuller account of shooting than the public NLL box score, which reports shots on goal but not a total combining goals, saves, misses, and blocks \citep{nll2026boxscore}. For team shots $S_g$ in game $g$,

\begin{equation}
\begin{aligned}
N_g &= N_g^{\mathrm{goal}}+N_g^{\mathrm{save}}+N_g^{\mathrm{miss}}+N_g^{\mathrm{block}}, &
\mathrm{OnTargetRate}_g &= \frac{N_g^{\mathrm{goal}}+N_g^{\mathrm{save}}}{N_g}, \\
\mathrm{xG}_g &= \sum_{i\in S_g}\mathrm{xG}_i, &
\mathrm{FAE}_g &= G_g-\mathrm{xG}_g.
\end{aligned}
\label{eq:game_level_summary_metrics}
\end{equation}

A low goal total with low xG indicates limited opportunity value, whereas a low goal total with substantial xG indicates finishing below expectation. Because game totals aggregate over many shots, however, $\mathrm{xG}_g$ tracks attempt volume closely, and the on-target rate and full attempt count carry much of the descriptive information the box score omits.

\paragraph{Passing and pick connections.}
Passer--shooter tables report linked-shot count, total xA, and average xA per linked shot, crediting the final passer whether or not the attempt is converted. Picker--shooter tables report qualifying pick count and on-ball and off-ball xPV. The latter must be read with the validation in Section~\ref{subsec:xpv_validation}: individual and pairwise xPV magnitudes are not distinguishable from model noise, so these tables identify frequently used two-player combinations for review rather than ranking their effectiveness. A single shot may generate both xA and xPV when an off-ball pick frees a teammate to receive a pass and shoot; the shared attribution across roles is intentional, and pairwise values depend on shared playing time and opportunity volume.

\paragraph{Automated reporting.}
Once a game is annotated and validated, the pipeline applies the model, aggregates team and player measures, and generates consistent tables and figures. Historical reports use out-of-fold probabilities so each shot is valued by a model that did not see its game; newly collected games use the deployable model fitted on the full development set. Automated checks reconcile shot and goal totals, verify attacking-direction normalization, and flag missing or invalid fields before a report is produced. High-danger classification is deliberately excluded because no xG threshold was prespecified, and the current probability distribution is too compressed to support one. Automation covers feature generation, prediction, aggregation, and presentation, but not video annotation, which remains the binding constraint on scaling the framework.

\section{Discussion}
\label{sec:discussion}

\subsection{Principal Findings and Implications}
\label{subsec:discussion_findings}

The results provide direct responses to the research questions introduced in Section~\ref{subsec:introduction_contributions}. First, the contextual baseline random forest had the lowest observed aggregate held-out probability error among the nine evaluated combinations, with out-of-fold log loss 0.4189 and Brier score 0.1260. The same Leave-One-Game-Out predictions were used both to compare specifications and to identify this lowest-loss model, so the reported performance is descriptive rather than an independent post-selection assessment. The more informative comparison is against the base-rate benchmark: the lowest-loss specification improved on a constant predictor by 1.22\% in log loss and 1.50\% in Brier score, and five of the nine specifications performed worse than that benchmark. The recorded shot context therefore supplies limited probability information beyond the goal rate at this sample size, and the ROC AUC of 0.604 indicates correspondingly weak discrimination. The ranking among specifications should be read as a comparison among models that all sit close to the base rate rather than as evidence that any of them captures shot quality with high fidelity. The observed advantage of random forest is specific to the fixed configurations and thirteen-game sample and does not establish general superiority for NLL xG.

Second, adding the two pick-related fields did not improve the primary probability metrics: log loss rose to 0.4195 with two-man-action role and 0.4197 with qualifying pick type, and Brier scores were marginally worse. The pick-aware model's ECE was lower and its ranking metrics were similar, illustrating why multiple metrics must be interpreted together. Statistical power offers a plausible explanation. One-hot encoding seventeen source fields creates a sparse design relative to the 151 observed goals, and the fifteen observed off-ball picks cannot support stable estimation of that category. The action fields may therefore describe meaningful tactics without adding recoverable predictive information at this sample size. Game-level results varied materially, with baseline log loss ranging from 0.295 to 0.597 across held-out games---a spread far larger than the aggregate differences among specifications. More broadly, these results demonstrate that plausible tactical annotations do not necessarily improve out-of-sample probability estimates when their categories are sparse and the number of independent games is limited. This is a useful negative result for subsequent event-collection efforts in the sport, which face the same tradeoff between annotation richness and sample size.

The third research question concerned extending shot probabilities to player evaluation. The completed annotations support xG for shooters and xA for recorded final passers; counterfactual xPV for qualifying pickers is retained as an exploratory model contrast. In the displayed exploratory augmented ranking, Ryan Lanchbury (No.~7) had the largest combined score through high-volume final-pass creation, Connor Fields (No.~10) combined the largest shooting workload with the largest total xG, and Thomas McConvey (No.~12) had the largest descriptive xPV total. Reporting xG per shot alongside total xG exposes the distinction between cumulative involvement and average opportunity quality. That distinction should nevertheless be applied cautiously here. Individual shot probabilities do vary substantially, spanning 0.028 to 0.470 with a median of 0.139, but player averages compress toward the base rate because each player takes a mixture of opportunities: among the nine players in Table~\ref{tab:player_oi_results}, xG per shot ranges only from 0.130 to 0.166. Total xG therefore closely tracks attempt volume, and core OI correspondingly reflects recorded involvement counts weighted by modest differences in average quality. The attribution framework is better understood as a structured accounting of recorded offensive roles than as a finely resolved measure of opportunity quality.

The validation diagnostics in Section~\ref{subsec:xpv_validation} sharpen what counterfactual xPV can support. Under a row-permutation diagnostic in which the pick fields are permuted jointly within each training fold, the mean absolute contrast was slightly larger than the observed value, 74 of 200 replicates produced a smaller mean absolute contrast than the real data, and a placebo intervention applied to no-pick shots produced contrasts of comparable size. The magnitude of xPV is accordingly indistinguishable from arbitrary partitioning, and neither individual values nor player totals should be compared as calibrated goal-equivalent quantities. Positive contrasts occurred for 82.2\% of qualifying picks against a null mean of 55.1\%, exceeding all 200 permutation replicates, and every one of the thirteen held-out games produced positive contrasts for more than half its qualifying picks. This is a descriptive model-response pattern, not a formal significance test: 200 replicates provide limited tail resolution, and row-wise shuffling does not preserve the composition of pick events within games. The pattern also does not make xPV a player-discriminating measure: because picker identity is never a model feature, player xPV totals correlate with raw pick count at Spearman $\rho=0.98$ and are close to a rescaled volume statistic. The contribution of xPV in this study is accordingly the counterfactual definition and its diagnostic framework rather than the specific values reported. Because the counterfactual holds observed shot location fixed, it also omits any value operating through the pick producing a better location in the first place, which a possession-level formulation would capture.

The framework should complement rather than replace conventional statistics and video analysis. It identifies where opportunities occurred and which players were associated with their creation, but it does not establish that a particular tactic or action caused the resulting scoring probability. Coaches and analysts can use the measures to locate relevant possessions and player combinations before returning to video to evaluate defensive pressure, movement, decision-making, and other factors not represented by the model. Given the modest margin over the base rate, that supporting role is the appropriate one for the framework in its current form.

\subsection{Limitations}
\label{subsec:discussion_limitations}

Several limitations affect the interpretation and generalizability of the findings. The dataset contains 1,006 shot attempts and 151 goals from thirteen Rochester Knighthawks games. Although Leave-One-Game-Out cross-validation preserved the game-level structure, every test fold involved the same focal offensive team. The evaluation therefore measures generalization to unseen games within the case study rather than to different NLL teams, seasons, or styles of play. Independent league-wide validation is required before the model can be interpreted as a general estimate of NLL shot quality.

The dataset was also constructed through manual review of broadcast video. Recorded coordinates and contextual classifications depend on camera angle, visibility, event definitions, and observer judgment. Goals were checked against official statistics, each game was reviewed a second time, and automated validation was used to identify inconsistencies. These procedures reduce annotation error but do not replace independent annotation or a formal assessment of inter-rater reliability.

An illustrative coordinate-perturbation sensitivity scenario helps show the possible consequence of spatial imprecision. Shot locations are recorded by clicking a plotter, so repeated coding of the same attempt would place it within some distance of the original point. Adding Gaussian noise of a given standard deviation to both recorded axes, recomputing distance, angle, and the behind-goal indicator from the perturbed point, and rescoring each held-out game with the same fitted model gives the change in predicted xG under the assumed perturbation. At one foot of noise on each axis the mean absolute change in predicted xG was 0.0116, and at three feet it was 0.0227. For comparison, the out-of-fold xG values themselves have a standard deviation of 0.0717. Coordinate error of a few feet can therefore produce appreciable changes in individual shot probabilities, reinforcing the treatment of xG as a contextual event-level estimate rather than a precise measurement of any single attempt. These figures describe the chosen scenarios, not a bound or calibrated measurement-error model: real plotting error may be non-Gaussian, correlated across axes, and dependent on court region or camera angle. The perturbation analysis was not propagated through model refitting or the row-permutation diagnostic, so it does not establish robustness of the xPV results to coordinate error.

The lowest-loss general xG specification includes spatial, temporal, situational, passing-presence, and offensive-context fields, but it does not account for defensive pressure, goaltender positioning, screen quality, shot velocity, shooter movement, pass speed, or complete preceding sequences. The resulting probabilities should therefore be interpreted as contextual event-level estimates rather than complete measurements of shot quality.

The attribution metrics also use deliberately limited definitions. xA credits only the recorded final passer and only when the pass produces a shot. Counterfactual xPV credits a picker only when a qualifying on-ball or off-ball pick directly produces an attempt. Neither measure captures earlier actions, longer sequences, or advantages that do not lead immediately to a shot. xA assigns the resulting shot's xG to an additional participant, while xPV assigns the observed-minus-no-pick probability contrast. Both remain forms of shared role attribution rather than estimates of unique causal contribution. Core OI values should not be summed as conserved team scoring value because the same shot can credit both shooter and passer; the exploratory xPV extension adds a further nonconserved role attribution.

The three algorithms used fixed prespecified settings rather than exhaustive or nested hyperparameter optimization. Conclusions about algorithm performance are therefore limited to the evaluated configurations. Future work should use nested game-level tuning after the dataset contains enough independent games to support it.

OI and its components are also influenced by playing time, offensive role, and opportunity volume. Complete possession, shift, and time-on-floor data were unavailable, preventing per-possession, per-minute, or usage-adjusted comparisons. Per-event rate statistics partially address this limitation but remain variable in small player samples. FAE is particularly susceptible to short-term finishing variation and should not be interpreted as stable shooting ability without substantially more attempts.

Finally, the estimated relationships are predictive rather than causal. Unmeasured defensive and tactical conditions may influence both an action and its outcome. Uncertainty intervals for player xG, xA, xPV, and OI were not estimated; with thirteen game clusters, an appropriately clustered interval would in any case be too imprecise to separate the players reported here, and adding one is a priority once the sample grows. These constraints reinforce the study's position as an initial case study and documented analytical baseline rather than a definitive league-wide player-rating system.

\subsection{Future Work}
\label{subsec:discussion_future_work}

The planned next phase is to build a league-wide model by collecting additional observations beyond one team and thirteen games. Adding shots from other NLL teams, opponents, venues, and seasons would increase the diversity of attempts and provide larger samples for infrequently observed categories. A league-wide dataset would also permit external validation in which models are developed using one set of teams or seasons and evaluated on teams or seasons not involved in model selection. Performance and calibration should then be monitored over time to determine when recalibration is required.

Future data collection should assess annotation reliability. A subset of games could be independently recorded by multiple annotators, allowing inter-rater agreement to be measured for coordinates, outcomes, passes, and pick classifications. Formal rules and an adjudication process could then be refined for variables that produce inconsistent classifications. Computer-vision and automated event-detection methods may eventually reduce the required manual effort, although automated observations would still require validation against carefully reviewed events.

A larger sample would support renewed evaluation of contextual features and collection of information not represented in the current model, including defensive pressure, goaltender positioning, screens, shot velocity, shooter movement, pass direction, and preceding action sequences. Hierarchical models could estimate team, shooter, and goaltender effects while accounting for repeated observations involving the same participants. Flexible nonlinear and sequence-based models should be reconsidered only after the expanded sample can support their additional complexity.

The immediate player-attribution priority is to examine the stability of xG, xA, xPV, OI, and FAE across games and seasons and determine the samples needed for reliable player comparisons. Possession, shift, and time-on-floor data would enable usage-adjusted measures. More off-ball examples are especially important because only 15 such picks were observed. Future xPV diagnostics should use more permutation replicates and preserve game-level event composition. Future work should also compare the present conditional xPV contrast with possession-level approaches capable of estimating value mediated through improved shot location or a longer offensive sequence. Indirect or delayed pick effects should remain separate from the directly shot-producing definition.

Finally, the probability-based framework can be extended beyond offensive shot creation. Potential applications include goals saved above expected for goaltenders, defensive-impact measures, transition evaluation, and possession-value models. These extensions would support a more complete representation of professional box-lacrosse performance while preserving the transparent event definitions and game-level validation principles established in this study.

\section{Conclusion}
\label{sec:conclusion}

Traditional offensive statistics provide valuable summaries of professional box lacrosse performance but do not fully convey the quality of scoring opportunities or the different roles involved in creating them. Goals do not account for shot difficulty, assists depend on the receiving player converting the opportunity, and shots on goal exclude missed and blocked attempts. Players who contribute through off-ball picks may also create valuable opportunities without receiving credit in conventional box-score statistics. This study addressed these limitations by developing a statistical and machine-learning framework for quantifying recorded offensive impact in professional box lacrosse.

The empirical analysis used 1,006 manually annotated shot attempts, including 151 goals, from thirteen Rochester Knighthawks games played from January~3 through April~4, 2026. Each observation was processed through a validated feature-engineering pipeline, including correction of attacking direction before spatial features were calculated. Logistic regression, random forest, and extremely randomized trees were crossed with three nested feature specifications and evaluated using Leave-One-Game-Out cross-validation. The contextual baseline random forest had the lowest observed out-of-fold log loss (0.4189) and Brier score (0.1260) among the fixed specifications. Because the same cross-validation results were used for comparison and selection, these metrics are descriptive rather than an independent post-selection estimate. They improve on a constant base-rate predictor by 1.22\% and 1.50\% respectively, and five of the nine evaluated specifications failed to beat that benchmark at all. Adding two-man-action role and qualifying pick type did not improve the primary metrics. The framework's contribution therefore lies in the event representation, validation design, and attribution structure rather than in the predictive strength of the current model.

The resulting probabilities support core player attribution through xG for shooters and xA for recorded final passers. A completed annotation pass also enabled an exploratory counterfactual Expected Pick Value for qualifying on-ball and off-ball pickers. Connor Fields generated the largest total xG, while xG per shot distinguished average modeled opportunity quality from cumulative shooting value. The displayed xPV totals and augmented rankings remain descriptive: a row-permutation diagnostic showed that xPV magnitude was indistinguishable from model noise, while the share of positive contrasts exceeded all 200 replicates. The latter result has limited tail resolution and does not preserve game-level pick composition, so it is not presented as formal inferential evidence. Current xPV values should be treated as exploratory directional contrasts rather than validated measures of causal pick impact or goal-equivalent player value.

Because the dataset contains 1,006 shots and 151 goals from thirteen games involving one focal team, the model should be interpreted as an initial case study and documented methodological baseline rather than a definitive league-wide estimate of shot quality or player value. The planned next phase is to collect additional data and build a league-wide model, enabling independent team- or season-level validation and nested tuning on a larger set of games. The present study documents a transparent method for assigning supported recorded shot-producing value to shooters and final passers, and for examining qualifying pickers through a separate exploratory contrast, while keeping finishing and per-event rates visible as distinct dimensions.

\section*{Data Availability}
The manually annotated event data are not publicly available. The annotations were derived from NLL+ broadcast video, and the event file is being retained for potential proprietary research and practitioner applications. Variable definitions, sentinel encodings, and the feature dictionary are provided in Appendix~\ref{app:feature_dictionary} so that the schema can be implemented on an independently collected dataset. This study therefore contributes a documented methodology rather than an open dataset or a directly replicable numerical analysis.

\section*{Code Availability}
The analysis code is not publicly released and is being retained for potential proprietary use. Sections~\ref{sec:methodology} and~\ref{subsec:xpv_validation} and Appendix~\ref{app:hyperparameters} document the principal preprocessing, model-comparison, attribution, and diagnostic procedures to support independent methodological implementation. Exact replication of the reported numerical results requires the unreleased event data and code.

\section*{Research Independence, Funding, and Competing Interests}
The author conducted this study as an independent researcher. The Rochester Knighthawks did not sponsor, fund, supply data for, participate in, or review the analysis. Rochester was selected as the focal case because the research originated from a Rochester-specific coaching question and the video-annotation work was practically feasible. The study received no external funding, and the author declares no competing interests.

\section*{Acknowledgements}
The author thanks Stew Monture and Cory Bomberry for sharing the coaching concept of a \textit{pick assist}, which informed the proposed Expected Pick Value definition.

\bibliographystyle{plainnat}
\bibliography{references}

\begin{appendices}
\section{Current Feature Dictionary}
\label{app:feature_dictionary}

The frozen corrected dataset contains the raw event fields and derived variables listed below. Model roles refer to the default experiment; player identifiers remain available for aggregation but are not included unless an explicit optional player-identity analysis is requested. Picker number is never a model feature.

\footnotesize
\begin{longtable}{@{}>{\raggedright\arraybackslash}p{0.225\textwidth}>{\raggedright\arraybackslash}p{0.18\textwidth}>{\raggedright\arraybackslash}p{0.465\textwidth}@{}}
\caption{Variables in the corrected processed dataset.}
\label{tab:complete_feature_dictionary}\\
\toprule
\textbf{Variable} & \textbf{Role} & \textbf{Definition} \\
\midrule
\endfirsthead
\toprule
\textbf{Variable} & \textbf{Role} & \textbf{Definition} \\
\midrule
\endhead
\bottomrule
\endfoot

\texttt{source\_file}, \texttt{source\_row} & Audit & Original CSV filename and one-based source row. \\
\texttt{game\_id}, \texttt{shot\_id} & Grouping/identifier & Game key for LOGO folds and unique event identifier. \\
\texttt{Quarter} & Baseline numeric & Regulation quarter or overtime period, with overtime encoded as 5. \\
\texttt{Time In Qtr} & Raw & Display clock retained as entered. \\
\texttt{time\_remaining\_seconds} & Baseline numeric & Period clock converted to seconds, including fractional seconds. \\
\texttt{shot clock} & Raw/sentinel & Recorded shot clock; $-1$ denotes disabled. \\
\texttt{shot\_clock\_seconds} & Baseline numeric & Numeric shot clock with disabled values represented as zero. \\
\texttt{shot\_clock\_disabled} & Baseline numeric & Indicator equal to one when the raw shot clock is $-1$. \\
\texttt{StrengthState} & Baseline categorical & Recorded manpower state. \\
\texttt{Offense} & Baseline categorical & Settled or transition offense. \\
\texttt{swing} & Baseline categorical & Recorded if shot came 1 or 2 passes after a swing \\
\texttt{shottype} & Baseline categorical & Recorded release or finishing technique. \\
\texttt{off pass} & Baseline categorical & Recorded pass-versus-self-created state. \\
\texttt{Score diff} & Baseline categorical & Score-differential category from Rochester's perspective. \\
\texttt{Pulled goalie} & Baseline categorical & Recorded pulled-goalie state. \\
\texttt{two man action} & Extension categorical & Ball carrier, picker, or no two-man action; first intervention field for xPV. \\
\texttt{qualifying\_pick\_type} & Extension categorical & On-ball, off-ball, or no qualifying pick; second intervention field for xPV. \\
\texttt{shot result}, \texttt{is\_goal} & Outcome & Human-readable result and binary target. \\
\texttt{Shooter Number} & Attribution identifier & Shooter jersey number; excluded from the default model. \\
\texttt{passer number} & Attribution identifier & Final passer jersey number; $-1$ means none recorded. \\
\texttt{passer\_present} & Baseline numeric & Indicator derived from passer number without revealing identity. \\
\texttt{picker\_number} & Attribution identifier & Credited picker jersey number; $-1$ means no qualifying picker and the field is always excluded from prediction. \\
\texttt{qualifying\_pick} & Audit/aggregation & Indicator derived from qualifying pick type. \\
\texttt{X}, \texttt{Y} & Raw spatial & Original court coordinates. \\
\texttt{attacking\_goal\_x} & Preprocessing/audit & Inferred attacked-goal coordinate for the game-period group. \\
\texttt{X\_attack\_normalized} & Baseline numeric & Horizontal coordinate after mirroring left-goal periods. \\
\texttt{shot\_distance} & Baseline numeric & Euclidean distance to the normalized attacking goal. \\
\texttt{shot\_angle\_degrees} & Baseline numeric & Absolute angle to the normalized attacking goal. \\
\texttt{behind\_goal} & Baseline numeric & Indicator for attempts originating beyond the goal line. \\
\end{longtable}
\normalsize

\subsection{Feature Specifications}

The baseline combined ten numeric source fields and seven categorical source fields. The two-man specification added \texttt{two man action}. The full pick-aware specification added both \texttt{two man action} and \texttt{qualifying\_pick\_type}. One-hot encoding determines the final matrix width within each training fold, so source-field counts are reported instead of a single encoded-column total.

\section{Current Experiment Settings and Supplemental Results}
\label{app:hyperparameters}

\subsection{Common Settings}

Every combination used thirteen-fold Leave-One-Game-Out cross-validation. Numeric fields were median-imputed and standardized within the training fold. Categorical fields were most-frequent imputed and one-hot encoded with unknown held-out levels ignored. Aggregate out-of-fold log loss was the primary selection metric, followed by Brier score and ten-bin ECE as tie-breakers. ROC AUC and average precision were supporting ranking measures. Random seed 20251214 was used throughout.

\begin{table}[ht]
\centering
\small
\caption{Fixed estimator settings in the corrected experiment.}
\label{tab:current_model_hyperparameters}
\begin{tabular}{@{}p{0.22\textwidth}p{0.67\textwidth}@{}}
\toprule
\textbf{Model} & \textbf{Settings} \\
\midrule
Logistic regression & $C=1.0$; maximum 3,000 iterations; no class weighting; standardized numeric inputs \\
Random forest & 500 trees; minimum leaf size 5; \texttt{max\_features=sqrt}; no class weighting; parallel fitting \\
Extra Trees & 500 trees; minimum leaf size 5; \texttt{max\_features=sqrt}; no class weighting; parallel fitting \\
\bottomrule
\end{tabular}
\end{table}

\subsection{Game-Level Variation}

For the selected baseline random forest, held-out game log loss ranged from 0.294984 to 0.596580, with an unweighted game mean of 0.421055 and standard deviation 0.086494. The full pick-aware random forest ranged from 0.296498 to 0.600137 and had lower game-level log loss in five of thirteen games. Aggregate pooled metrics, rather than unweighted fold means, governed model ranking because games contained different numbers of shots.

\subsection{Software Environment}

The corrected rerun used Python 3.12.13, NumPy 2.5.2, pandas 2.2.3, SciPy 1.18.1, scikit-learn 1.7.2, and joblib 1.5.2. The public manuscript package does not include the processed dataset, analysis code, out-of-fold predictions, or fitted model objects. The versions and settings are reported to document the computational environment; exact numerical replication requires the unreleased research materials.

\section{Complete Attribution Roster}
\label{app:full_attribution}

Table~\ref{tab:full_attribution} lists every player who accumulated any xG, xA, or xPV in the annotated
sample, providing the full denominator for the descriptive augmented-score ranks reported in
Section~\ref{sec:empirical_oi}. Players are identified by jersey number. The nine players discussed in the
main text are those credited with at least five qualifying picks; all others were credited with at most one.
As noted in Section~\ref{subsec:oi_aggregation}, neither core OI nor the exploratory augmented score is
conserved across players, and the displayed sum should not be read as team scoring value.

\begin{table}[ht]
\centering
\small
\setlength{\tabcolsep}{4pt}
\caption{Complete attribution roster, ordered by exploratory augmented Offensive Impact. xPV magnitudes are not distinguishable
from model noise (Section~\ref{subsec:xpv_validation}) and are reported for completeness only; the ordering is not a validated player-value ranking.}
\label{tab:full_attribution}
\begin{tabular}{@{}rrrrrrrrrr@{}}
\toprule
\textbf{Rank} & \textbf{No.} & \textbf{Shots} & \textbf{Goals} & \textbf{xG/shot} & \textbf{xG} &
\textbf{xA} & \textbf{Picks} & \textbf{xPV} & $\mathbf{OI^{expl}}$ \\
\midrule
1 & 7 & 131 & 22 & 0.144 & 18.83 & 45.27 & 7 & 0.211 & 64.32 \\
2 & 10 & 233 & 31 & 0.140 & 32.56 & 27.59 & 6 & 0.154 & 60.30 \\
3 & 43 & 194 & 27 & 0.160 & 31.12 & 12.46 & 42 & 0.756 & 44.34 \\
4 & 12 & 165 & 24 & 0.161 & 26.57 & 10.91 & 60 & 1.159 & 38.64 \\
5 & 28 & 93 & 18 & 0.156 & 14.47 & 6.58 & 35 & 0.709 & 21.76 \\
6 & 42 & 43 & 4 & 0.144 & 6.18 & 4.39 & 39 & 0.622 & 11.20 \\
7 & 16 & 45 & 6 & 0.141 & 6.35 & 3.69 & 17 & 0.265 & 10.31 \\
8 & 81 & 18 & 4 & 0.213 & 3.84 & 1.90 & 0 & 0.000 & 5.73 \\
9 & 32 & 6 & 1 & 0.311 & 1.86 & 3.25 & 1 & -0.005 & 5.11 \\
10 & 26 & 17 & 2 & 0.196 & 3.34 & 1.73 & 0 & 0.000 & 5.07 \\
11 & 44 & 7 & 0 & 0.221 & 1.54 & 2.97 & 0 & 0.000 & 4.51 \\
12 & 88 & 12 & 6 & 0.210 & 2.53 & 1.77 & 0 & 0.000 & 4.29 \\
13 & 20 & 15 & 0 & 0.166 & 2.49 & 0.90 & 11 & 0.273 & 3.66 \\
14 & 11 & 9 & 3 & 0.130 & 1.17 & 1.52 & 5 & 0.105 & 2.80 \\
15 & 37 & 2 & 0 & 0.064 & 0.13 & 2.61 & 0 & 0.000 & 2.74 \\
16 & 74 & 7 & 2 & 0.216 & 1.51 & 0.39 & 0 & 0.000 & 1.91 \\
17 & 6 & 6 & 0 & 0.222 & 1.33 & 0.35 & 0 & 0.000 & 1.68 \\
18 & 54 & 1 & 1 & 0.080 & 0.08 & 0.92 & 1 & 0.012 & 1.01 \\
19 & 49 & 0 & 0 & --- & 0.00 & 1.00 & 0 & 0.000 & 1.00 \\
20 & 15 & 0 & 0 & --- & 0.00 & 0.59 & 0 & 0.000 & 0.59 \\
21 & 1 & 0 & 0 & --- & 0.00 & 0.49 & 0 & 0.000 & 0.49 \\
22 & 17 & 1 & 0 & 0.140 & 0.14 & 0.20 & 0 & 0.000 & 0.34 \\
23 & 29 & 0 & 0 & --- & 0.00 & 0.30 & 1 & 0.005 & 0.31 \\
24 & 13 & 1 & 0 & 0.290 & 0.29 & 0.00 & 0 & 0.000 & 0.29 \\
25 & 36 & 0 & 0 & --- & 0.00 & 0.13 & 0 & 0.000 & 0.13 \\
\midrule
\multicolumn{9}{@{}l}{\textit{Total}} & 292.52 \\
\bottomrule
\end{tabular}
\end{table}

\end{appendices}

\end{document}